\documentclass[arxiv]{melba}
\usepackage{dsfont}
\usepackage{graphics}
\usepackage{adjustbox}

\usepackage{mwe} 
\usepackage{subfig}
\usepackage{wrapfig}

\usepackage{soul}
\usepackage{amsmath,amsfonts}

\melbaid{2025:009}  
\doi{https://doi.org/10.59275/j.melba.2025-g137}
\melbaauthors{Xu and Geras}  
\email{yx2105@nyu.edu}
\volume{3}
\firstpageno{152}  
\melbayear{2025}  
\datesubmitted{06/2024}  
\datepublished{05/2025}  

\melbaspecialissue{Medical Imaging with Deep Learning (MIDL) 2020}
\melbaspecialissueeditors{Marleen de Bruijne, Tal Arbel, Ismail Ben Ayed, Hervé Lombaert}

\ShortHeadings{Understanding differences in applying DETR to natural and medical images}{Xu et al.}

\title{Understanding differences in applying DETR to natural and medical images}

\author{
	\firstname Yanqi \surname Xu\aff{1}\orcid{0009-0006-0744-4183},
	\name Yiqiu Shen\aff{2}\orcid{0000-0002-7726-2514},
    \name Carlos Fernandez-Granda\aff{1}\orcid{0000-0001-7039-8606},
    \name Laura Heacock\aff{2}\orcid{0000-0002-0900-0459},
    \name Krzysztof J. Geras\aff{2}\orcid{0000-0003-0549-1446}}
\affiliations{
	\num 1 \addr Center for Data Science, New York Univeristy, New York, NY, USA \\
	\num 2 \addr NYU Grossman School of Medicine, New York, NY, USA 
}

\newcommand{\new}[1]{\textcolor{black}{#1}}
\newcommand{\newer}[1]{\textcolor{black}{#1}}

\abstract{
Natural images depict real-world scenes such as landscapes, animals, and everyday items. Transformer-based detectors, such as the Detection Transformer, have demonstrated strong object detection performance on natural image datasets.  These models are typically optimized through complex engineering strategies tailored to the characteristics of natural scenes. However, medical imaging presents unique challenges, such as high resolutions, smaller and fewer regions of interest, and subtle inter-class differences, which differ significantly from natural images. In this study, we evaluated the effectiveness of common design choices in transformer-based detectors when applied to medical imaging. Using two representative datasets, a mammography dataset and a chest CT dataset, we showed that common design choices proposed for natural images, including complex encoder architectures, multi-scale feature fusion, query initialization, and iterative bounding box refinement, fail to improve and can even be detrimental to the object detection performance. In contrast, simpler and shallower architectures often achieve equal or superior results with less computational cost. These findings highlight that standard design practices need to be reconsidered when adapting transformer models to medical imaging, and suggest that simplicity may be more effective than added complexity in this domain. Our model code and weights are publicly available at \url{https://github.com/nyukat/Mammo-DETR}.}

\keywords{
Deep Learning, Machine Learning, Vision Transformers, Object Detection, Breast Cancer}

\begin{document}

\twocolumn[\maketitle]

\section{Introduction}

\begin{figure*}[t]
		\centering
		\includegraphics[width=0.8\linewidth, height=.2\textheight]{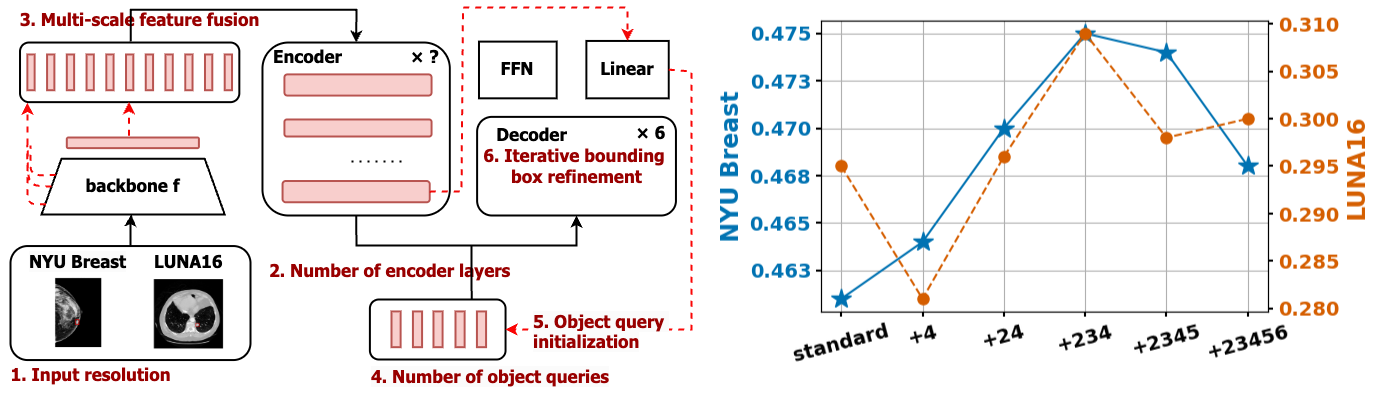}
		\caption{\textbf{An overview of the study}. In this work, we investigate key design choices in Deformable DETR using the NYU Breast Cancer Screening Dataset (NYU Breast) and LUNA16 Dataset. Specifically, we evaluate six design factors (highlighted in red): (1) input resolution, (2) number of encoder layers, (3) use of multi-layer feature fusion, (4) number of object queries, (5) query initialization method, and (6) use of iterative bounding box refinement. The graph on the right shows changes in Average Precision (AP) resulting from these design choices. \texttt{+2} indicates a reduction in encoder layers; \texttt{+3} removes multi-layer fusion; \texttt{+4} reduces object queries; \texttt{+5} adds query initialization; and \texttt{+6} enables iterative box refinement. Our findings suggest that a simplified architecture (\texttt{+2,3,4}) is better suited for medical datasets, leading to improved performance. }
    \label{fig:overview}
	\end{figure*}

Recent advances in computer vision have increasingly turned to transformer architectures \citep{vaswani2017attention} for tasks such as image classification and object detection \citep{dosovitskiy2020image, liu2021swin, carion2020end, touvron2021training}. With their inherent self-attention mechanisms, transformers effectively capture global dependencies and understand contextual relations across the entire image. These strengths have made transformer-based models a popular choice in natural image analysis. Their application in medical imaging has shown promising results, suggesting strong potential in this field as well \citep{chen2021transunet,dai2021transmed,valanarasu2021medical,zheng2022graph}.

Object detection is crucial in medical image analysis, as detection models identify the locations of abnormalities, which are important for medical diagnosis. Among transformer-based detectors, Detection Transformer (DETR) \citep{carion2020end} has gained popularity for its end-to-end training pipeline and elimination of non-differentiable post-processing steps such as Non-Maximum Suppression (NMS) \citep{girshick2014rich}. By leveraging the transformer architecture and directly optimizing the objective function, DETR achieves state-of-the-art results on natural image benchmarks such as MS COCO \citep{zhu2020deformable,zhang2022dino,zong2023detrs}. Its success has drawn intense research interest, leading to a range of highly engineered DETR variants aimed at boosting accuracy and training efficiency \citep{zhu2020deformable,zhang2022dino,chen2022conditional,wang2022anchor,chen2022group}. 

\begin{figure*}[h!]
    \centering
    \begin{minipage}{0.35\textwidth}
        \centering
    \includegraphics[width=\linewidth]{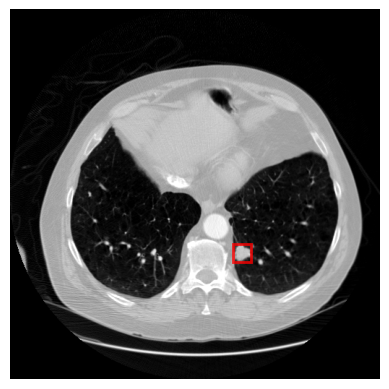} 
    \text{(a)}
    \end{minipage}
    \hfill
    \begin{minipage}{0.31\textwidth}
        \centering
      \includegraphics[width=\textwidth,height =1.25\textwidth]{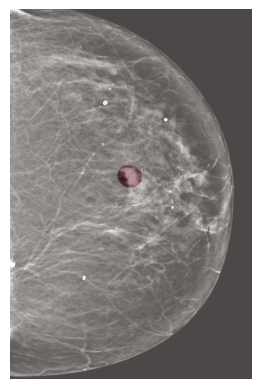}
       \text{(b)}
     \end{minipage}
    \begin{minipage}{0.31\textwidth}
        \centering
        \includegraphics[width=\textwidth,height = 0.6\textwidth]{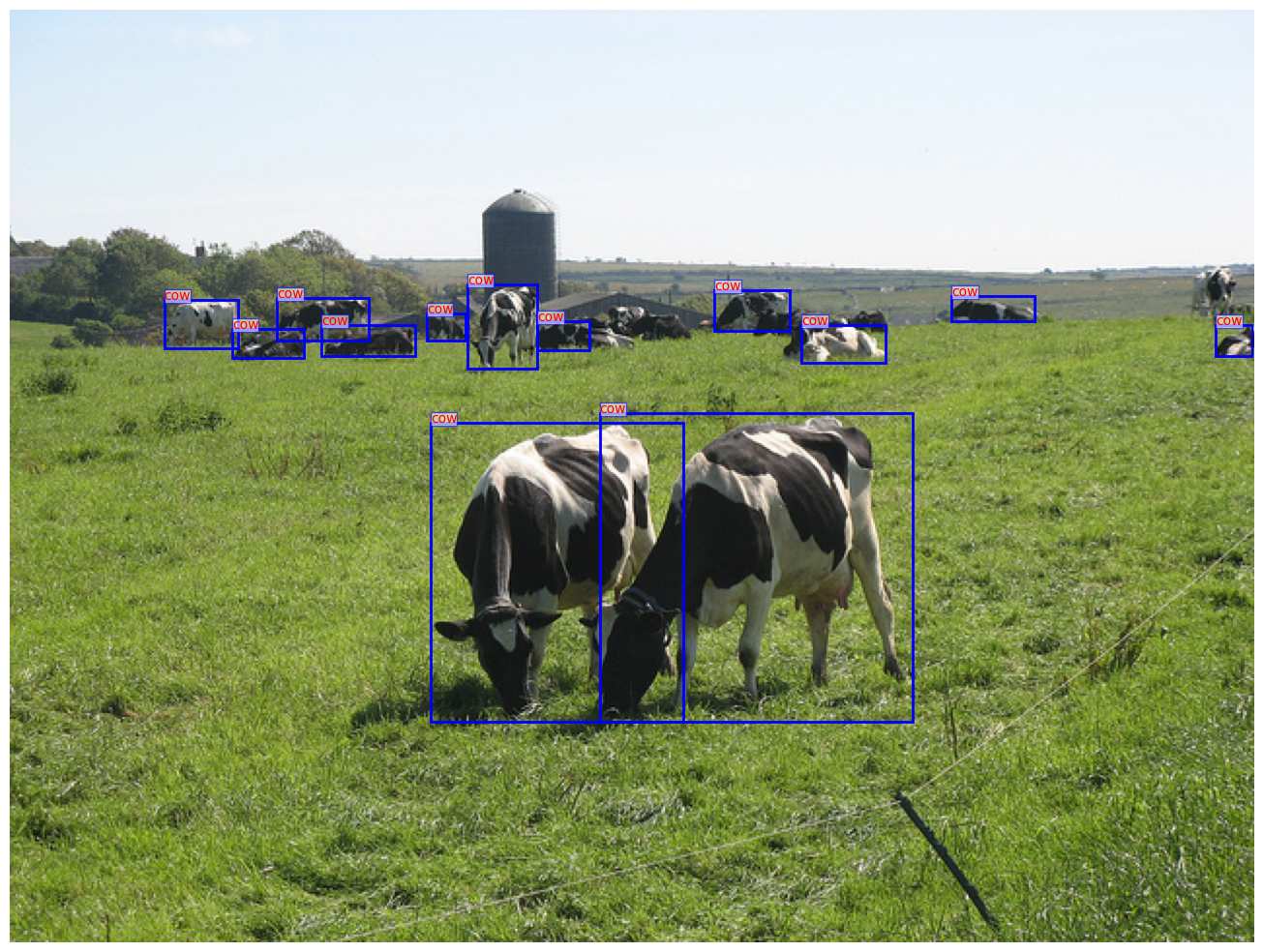}
        \centering
        \includegraphics[width=\textwidth,height = 0.6\textwidth]{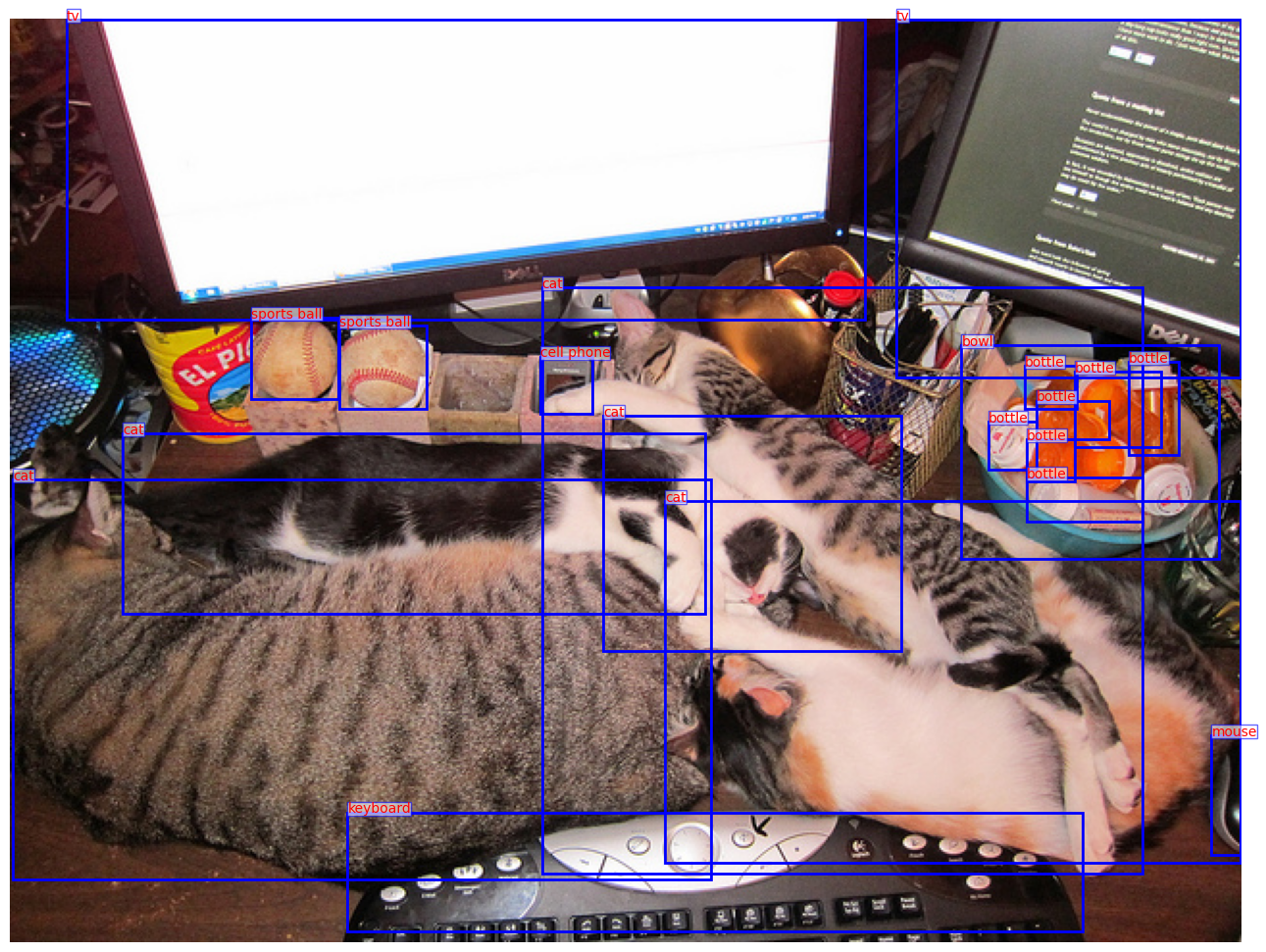}
         \text{(c)}
    \end{minipage}
    \caption{\newer{\textbf{Example images from the LUNA16, NYU Breast Cancer Screening, and MS COCO datasets.} (a) A chest CT slice from LUNA16 showing a lung nodule (red box). Among images that contain nodules, the average number of objects per image is 1.21. This dataset has one class.
(b) A mammogram from the NYU Breast Cancer Screening dataset showing a cancerous lesion (red highlight). In images containing lesions, there are on average 1.10 objects per image. This dataset has two classes.
(c) Two example images from the MS COCO dataset. These illustrate the complexity of natural scenes, with multiple overlapping objects of varying sizes. On average, MS COCO images contain 7.33 objects per image. MS COCO has 80 object classes.}}
    \label{fig:image-examples}
\end{figure*}

Despite the success of DETR architectures on natural image benchmarks, their direct application to medical imaging remains challenging due to fundamental differences between the two domains (Figure \ref{fig:image-examples}):
\begin{itemize}
    \item \textbf{High resolution and small regions of interest:} Medical images are often extremely high-resolution, with clinically relevant features, such as lesions or calcifications, occupying only small portions of the image \citep{moawad2023brain,heath2001digital}.
    \item \textbf{Standardized acquisition protocols:} Unlike natural images which have diverse backgrounds, medical images are acquired under standardized procedures, resulting in consistent anatomical structures and minimal background variability.
    \item \textbf{Few objects per image:} Medical images usually focus on a narrow range of abnormalities, resulting in fewer objects of interest and a narrower class space compared to the rich and diverse class space of natural images. Additionally, many medical images may not contain any objects at all.
    \item \textbf{Small and imbalanced data sets:} Medical imaging data sets are often small and exhibit a more unbalanced class distribution, as positive cases (i.e., unhealthy subjects) are usually much less common than negative cases (i.e., healthy subjects)\citep{galdran2021balanced,heath2001digital,wang2017chestx}.
\end{itemize} 

DETR-family models, such as Deformable DETR~\citep{zhu2020deformable}, incorporate complex design choices such as multi-scale feature fusion and iterative bounding box refinement to address challenges in natural image detection. However, their effectiveness in medical imaging is unclear, as the domain presents distinct characteristics: high resolution, small lesion size, limited object diversity, and class imbalance, which differ markedly from natural images. In such settings, detecting subtle features precisely is often more important than modeling diverse object scales or dense scenes. As a result, these complex design choices may introduce unnecessary computational overhead and memory cost without yielding performance gains.

In this study, we examine how DETR can be adapted to better suit medical imaging tasks. We hypothesize that a simplified model, tailored to the specific characteristics of medical data, can achieve comparable performance with reduced computational cost. \newer{To evaluate this, we use Deformable DETR~\citep{zhu2020deformable} as a baseline on two medical imaging datasets: the NYU Breast Cancer Screening Dataset~\citep{wu2019nyu} and the LUNA16 dataset~\citep{setio2017validation}, a public chest CT dataset focused on lung nodule detection (Figure \ref{fig:image-examples}). Those two datasets highlight the distinct features of medical images, such as high resolution, small lesions, and class imbalance.}

\new{Our experiments demonstrate that simplified DETR configurations—using fewer encoder layers, a single feature map, and no decoding enhancements—achieve detection performance on par with, or better than, standard Deformable DETR, while substantially reducing computational cost. These findings validate our hypothesis and highlight the potential of lightweight DETR variants as efficient and effective baselines for medical imaging.} The key findings of our work are:
\begin{itemize}
    \item Models with a reduced number of encoder layers and no multi-scale feature fusion learn faster without compromising detection performance. These changes maintain performance within $1\%$ in $\mathrm{AP}_{10,50}$ on both datasets, while accelerating training by up to 40\%.
    \item Increasing the number of object queries to around 100 queries improves localization and detection performance. Beyond this point, performance declines, primarily due to a rise in false positives that obscure true positive detections. 
    \item Decoding techniques such as object query initialization and iterative bounding box refinement, while beneficial for natural image detection, do not improve performance on medical datasets. In some cases, they degrade performance (e.g., a 0.7\% drop in $\mathrm{AP}_{10,50}$ for NYU Breast and 1.8\% drop for LUNA16), likely due to overfitting and limited positive examples.
\end{itemize}

\section{Background on DETRs}
DETR\new{~\citep{carion2020end}} offers several advantages over traditional detection models such as Mask-RCNN~\citep{he2017mask} and YOLO~\citep{redmon2016you}. Its transformer-based architecture enables more expressive feature representations, and its end-to-end training simplifies optimization and improves performance. However, DETR suffers from slow learning. To address this issue, various extensions have been proposed to accelerate training and improve detection performance~\citep{zhu2020deformable, wang2022anchor,chen2022conditional,zhang2022dino}. Deformable DETR\new{~\citep{zhu2020deformable}} stands out for its competitive performance on the MS COCO dataset~\citep{lin2014microsoft}. It introduces a deformable attention module that reduces training time by a factor of 10 and enables multi-scale feature fusion that improves detection, especially for small objects. Given its strong performance and widespread adoption in subsequent research~\citep{roh2021sparse,dai2021dynamic,zhang2022dino,yao2021efficient}, we adopt Deformable DETR as the baseline for our experiments. This section outlines the key components of DETR and Deformable DETR architectures.

\begin{figure}[h]
    \centering
    \subfloat[\centering vanilla DETR]{{\includegraphics[width=0.8\linewidth]{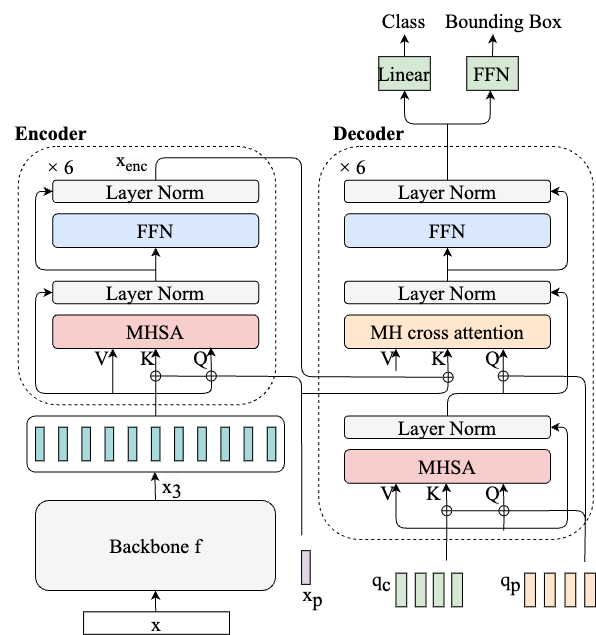} }}%
    \hspace{.03\textwidth}
    \subfloat[\centering Deformable DETR]{{\includegraphics[width=0.8\linewidth]{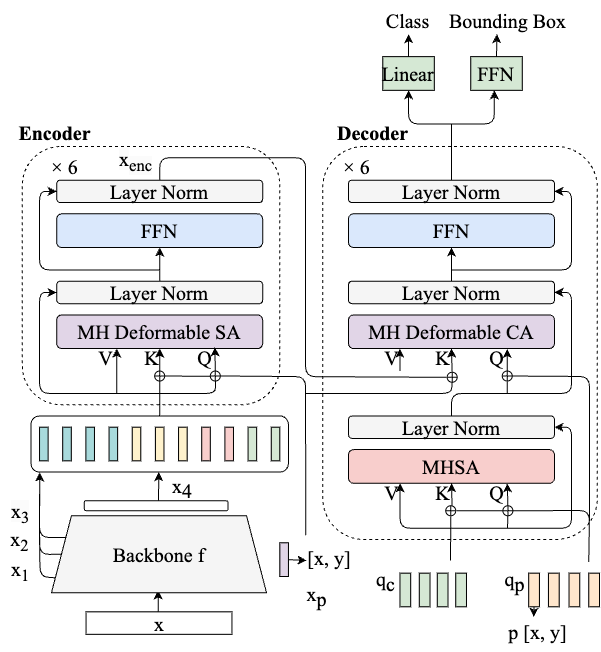} }}%
    \caption{\textbf{Architecture of DETR and Deformable DETR.} Both architectures consist of a backbone, an encoder-decoder transformer and a prediction head. Deformable DETR differs from DETR in its utilization of multiple feature level fusions and its application of deformable attention instead of standard attention mechanism. \new{Abbreviations: MHSA: multi-head self-attention module; MH cross attention: multi-head cross-attention; MH Deformable SA: multi-head deformable self-attention module; MH Deformable CA: multi-head deformable cross-attention module; FFN: feedforward network.}}
    \label{fig:DETR}
\end{figure}

\paragraph{DETR} 
DETR\new{~\citep{carion2020end}} consists of a backbone, an encoder-decoder transformer, and a prediction head, as illustrated in Figure~\ref{fig:DETR}(a).

Given an input image $x \in \mathbb{R}^{C_0\times W_0\times H_0}$, \new{where $C_0$ is the number of channels and $W_0$ and $H_0$ are the width and height}, the backbone network $f$ produces a low-resolution activation map $x_s= f(x) \in \mathbb{R}^{C\times W\times H}$, \new{with $C$ significantly larger than $C_0$. \new{The specific sizes of $W,H$ and $C$ depend on the choice of backbones. For instance, when using Swin-T \citep{liu2021swin} as the backbone}, the spatial dimensions are downsampled to $W=W_{0}/32$ and $H=H_{0}32$, and the number of channels is increased to $C=768$.} This map is further processed by a $1\times1$ convolution to collapse the channel dimension $C$ into a smaller size $d$, resulting in image tokens $x_f = \mathrm{conv}(x_s)\in \mathbb{R}^{WH \times d}$. To preserve spatial information in the original image, each token is paired with a positional encoding, denoted by $x_p\in \mathbb{R}^{WH \times d}$. The encoder is a standard attention-based transformer where each layer consists of a multi-head self-attention module (MHSA) followed by a feedforward network (FFN). For an in-depth formalization of MHSA, refer to the Appendix \ref{app:mhsa}. Typically, the DETR encoder consists of 6 layers. The encoder preserves the dimension of the input, producing $x_{enc}\in \mathbb{R}^{WH \times d}$.

The decoder receives two inputs, the encoded features $x_{enc}$ and $N$ object queries $q \in\mathbb{R}^{N\times d}$. Object queries play a central role in the DETR architecture. They are learnable embeddings that work as placeholders for potential objects in an image. Each of them attends to the specific regions of the image and is individually decoded into a bounding box prediction. Each object query is the sum of two learnable embeddings: content embeddings $q_c\in\mathbb{R}^{N\times d}$, initialized as zero vectors, and positional embeddings $q_p\in\mathbb{R}^{N\times d}$, indicating each query's position. More methods for initializing object queries are discussed in Section \ref{sec:design-choices}. Decoder layers consists of a MHSA, enabling inter-query learning, and multi-head (MH) cross-attention to integrate encoder features, and an FFN. The formalization of MH cross-attention is detailed in the Appendix \ref{app:mh-cross}.

After the decoder, each object query is independently decoded into bounding box coordinates and class scores through a three-layer FFN and a linear layer respectively.

\paragraph{Deformable DETR}
Deformable DETR\new{~\citep{zhu2020deformable}} improves upon DETR by introducing a deformable attention module, which accelerates training and enhances the detection of small objects. The architecture of Deformable DETR is illustrated in Figure~\ref{fig:DETR}(b).

Unlike the standard attention mechanism that calculates attention scores between all query-key pairs, resulting in $(WH)^2$ pairs for a feature map of size $W\times H$, deformable attention selectively computes attention scores on a subset of $k << WH$ keys for each query. The subset is selected through a learnable key sampling function, allowing the model to focus on the most informative regions for each query. For a detailed formalization of the deformable attention module, refer to Appendix \ref{app:deformable-mhsa}.

For dense prediction tasks such as object detection, incorporating higher-resolution feature maps can substantially improve detection performance, especially for smaller objects \citep{he2017mask}. However, the complexity of the standard attention mechanism is quadratic with respect to the number of tokens, making it infeasible for multiple scales of feature maps. The deformable attention mechanism enables effective multi-scale feature fusion. Specifically, the encoder receives the output feature maps $x_1,x_2, x_3$ from the last three layers of the backbone, and a convolutional layer generates the lowest resolution feature map $x_4$. All four feature maps undergo a $1\times 1$ convolution and then are reshaped into a sequence of feature vectors of dimension $d$, denoted by $x_f\in \mathbb{R}^{M\times d}$. Each token is associated with a positional embedding, as well as a layer embedding to identify feature map level. Section \ref{sec:design-choices} explores the benefits of multi-feature fusion for medical imaging datasets.

Moreover, Deformable DETR introduces reference points in the deformable attention module. In the encoder, each query $q$ is associated with a 2D reference point $p_q=[x,y]$, denoting its location on the feature map. The key sampling function generates $k$ sampling offsets with respect to the reference point, and thus determines the $k$ keys for the query. 
Similarly in the decoder, the reference point of each object query $q$ is defined by a linear projection of its positional embedding $q_p$. In this way, each object query can be mapped to a position on the feature map. This approach allows object queries to focus on specific regions, significantly accelerating learning\new{~\citep{zhu2020deformable}}.

\paragraph{DETR in Medical Imaging}
DETR-based architectures have been widely applied to various medical imaging tasks, often with architectural tweaks to improve overall performance. For example, \cite{mathai2022lymph} leveraged a bounding box fusion technique in DETR to reduce the false positive rate in lymph nodes detection. MyopiaDETR~\citep{li2023myopiadetr} utilizes a Feature Pyramid Network to improve the detection of small objects in lesion detection of pathological myopia. COTR~\citep{shen2021cotr} embeds convolutional layers into DETR encoders to accelerate learning in polyp detection. Although these works achieved good performances, our experiments \new{indicate} that, contrary to the common understanding, simplifying the DETR architecture can improve accuracy and accelerate training. We identified a work that also points in this direction, Cell-DETR~\citep{prangemeier2020attention}, also reduces the number of parameters \new{tenfold}, achieving faster inference speeds while maintaining performance on par with state-of-the-art baselines. Finally, \cite{article} applied out-of-the-box Deformable DETR on mammography for mass detection. However, their focus is the effect of a data augmentation method on its detection performance. Despite these advances, a systematic exploration of the effectiveness and relevance of foundational DETR design choices remains \new{underexplored}.

\section{Methods}
\label{sec:design-choices}

\subsection{Design Choices}
In this section, we outline key design choices in Deformable DETR that are relevant to the unique characteristics of medical images: input resolution, the number of encoder layers, multi-scale feature fusion, the number of object queries, and two techniques enhancing the decoding process, query initialization and iterative bounding box refinement (IBBR). We investigated whether these components, which improve performance on natural image datasets, offer similar benefits when applied to medical imaging tasks.

\paragraph{Input resolutions}
Downsampling is commonly used in detection models to reduce computational cost and satisfy memory constraints. Natural images can be significantly downsized to $224 \times 224$ or $256 \times 256$ pixels without losing important features, such as edges, shapes, and textures that are necessary for accurate predictions. In contrast, medical images are often an order of magnitude larger. For example, X-ray images can reach up to $2500 \times 3056$ pixels~\citep{johnson2019mimic}, and CT scans are typically $512 \times 512$ pixels~\citep{setio2017validation}. These high-resolution medical images contain fine-grained details, such as small lesions or slight changes in tissue density, which are crucial for an accurate diagnosis \citep{sabottke2020effect, thambawita2021impact}. However, processing high-resolution medical images is often infeasible due to the high computational requirements. To address this trade-off, we evaluate performance across input resolutions ranging from 25\% to 100\% of the original size, aiming to identify the optimal input resolution that balances model accuracy with computational efficiency and memory usage.

\paragraph{Encoder complexity}
Medical imaging datasets differ from natural image datasets in several important ways. First, they are typically smaller due to limited patient availability. Second, images within a dataset tend to be homogeneous, focusing on a single body part, such as the brain, breast, or chest, with uniform grayscale textures (Figure \ref{fig:image-examples}). Third, while natural images contain hundreds or thousands of object classes, medical image datasets usually have far fewer object classes. For example, NIH Chest X-ray contains 14 classes~\citep{wang2017chestx}, DDSM has 2~\citep{heath2001digital}, and BraTS has 4~\citep{moawad2023brain}. As a result, there is much less variation in the data that the model has to capture. Given the principle that model complexity should align with task complexity~\citep{geman1992neural}, we suspect that simpler, shallower architectures might be more appropriate for medical image analysis, helping mitigate overfitting and improve training efficiency. In addition, object sizes in medical images are typically more uniform than in natural scenes. For example, the standard deviation of normalized object sizes\footnote{Normalized object size refers to the area of the bounding box divided by the total image area.} is 0.025 in the NYU Breast Cancer Screening Dataset and 0.001 in LUNA16, compared to 0.16 in MS COCO. This raises questions about the use of multi-scale feature fusion in this domain, a technique primarily intended to improve detection across diverse object sizes. To investigate these hypotheses, we experimented with modifications to the encoder of Deformable DETR, including reducing the number of encoder layers and utilizing fewer scales of feature maps from the backbone. 

\paragraph{Number of object queries}
In DETR, each object query is individually decoded into a bounding box prediction. Thus, the total number of object queries determines how many objects the model can detect per image~\citep{carion2020end}. Most DETR models are optimized for natural image datasets such as MS COCO, where a single image can contain up to $100$ objects. Consequently, the number of object queries is usually set to $300$ in DETR models. In contrast, medical images rarely contain more than 10 objects, and most have only one or none. As a result, the default number of object queries used in standard DETR implementations may be excessive for medical applications, potentially leading to unnecessary computation or degraded performance. We therefore examine how reducing the number of object queries affects detection accuracy and efficiency on medical image datasets.

\paragraph{Decoding techniques}
Many DETR variants apply object queries initialization and iterative bounding box refinement (IBBR) to improve query decoding and increase detection accuracy~\citep{zhu2020deformable,zhang2022dino,yao2021efficient}. These methods have proven effective in boosting detection performance on natural image datasets, increasing average precision by 2.4 on the MS COCO dataset~\citep{zhu2020deformable}. In this study, we evaluate their effectiveness in the medical imaging domain. We tested three initialization strategies for the positional and content embeddings of object queries, as characterized by \cite{zhang2022dino}.
\begin{itemize}
    \item \textbf{Static queries}
    Both positional and content embeddings are randomly initialized as learnable embeddings. This offers maximum flexibility, but requires the model to learn where objects are likely located and what features represent those objects from scratch, potentially slowing convergence. Standard Deformable DETR uses this approach.
    \item \textbf{Pure query selection} Both content and positional embeddings are initialized from selected encoder features. Following~\cite{zhu2020deformable}, we apply the prediction head to the encoder output to select the top-$K$ features. Some other works use a regional proposal network \citep{yao2021efficient,chen2022conditional}. This leverages encoder knowledge to guide object queries and significantly accelerates training.
    \item \textbf{Mixed Query Selection:} Positional embeddings are initialized from encoder features (as above), while content embeddings remain randomly initialized. This hybrid strategy informs about the likely positions of objects through spatial priors while retaining flexibility in learning content representations from scratch. DETR with Improved DeNoising Anchor Boxes (DINO)~\citep{zhang2022dino} found that this method yields the best performance.

\end{itemize}

IBBR, first introduced in Deformable DETR, iteratively updates the reference points of object queries towards the objects of interest in each image. These reference points guide the deformable attention toward relevant regions to search for objects. Initially, they are randomly distributed across the image, ensuring broad coverage without any prior knowledge about where objects might be located. With IBBR, these reference points can move progressively towards the objects through each decoder layer, providing more accurate signals for attention. This technique has been extensively applied in subsequent DETR variants~\citep{zhu2020deformable,chen2022conditional,wang2022anchor, liu2022dab} and has been shown to effectively speed up training and improve detection performance. A detailed formalization is provided in the Appendix~\ref{app:refine}.

\subsection{Data and task}

\begin{figure}[h!]
    \centering
        \includegraphics[width=0.4
\textwidth,trim={0 0 0 0},clip]{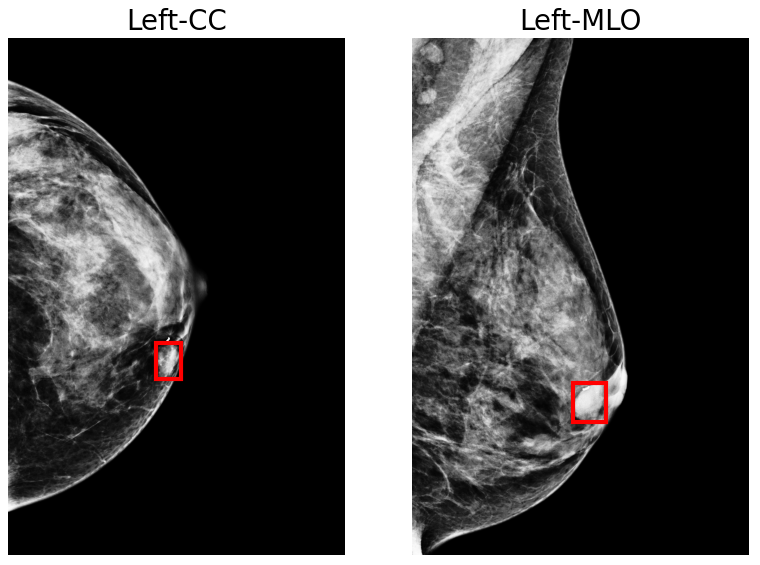}
\includegraphics[width=0.4
\textwidth,trim={0 0 0 0},clip]{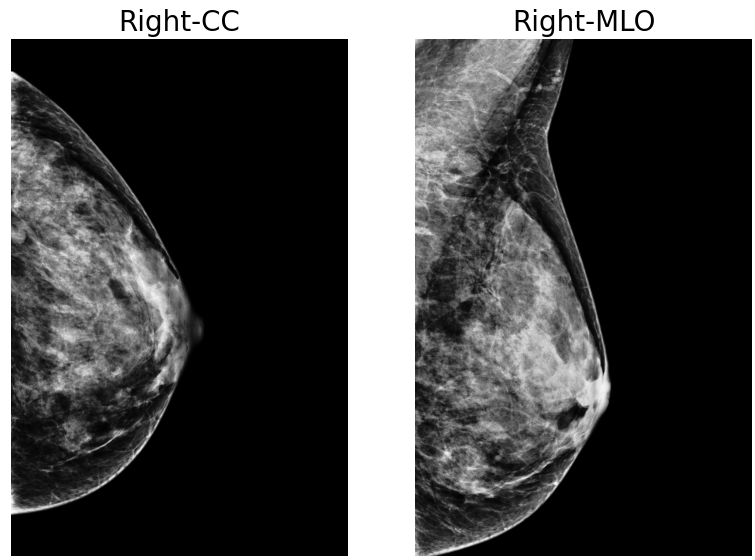}
    \caption{\textbf{An example screening mammography exam.} From left to right: left craniocaudal view \new{(L-CC)}, left mediolateral oblique view \new{(L-MLO)}, right craniocaudal view  \new{(R-CC)}, right mediolateral oblique view \new{(R-MLO)}. This patient has a benign lesion in the left breast. It is marked with a red bounding box from both views of the left breast.}
    \label{fig:mammo}
\end{figure}

\paragraph{NYU Breast Cancer Screening Dataset (NYU Breast)}~\citep{wu2019nyu} contains $229,426$ digital screening mammography exams from $141,472$ patients screened at NYU Langone Health. Each exam includes a minimum of four images, each with a resolution of $2944 \times 1920$, covering two standard screening views: craniocaudal (CC) and mediolateral oblique (MLO), for both the left and right breasts. An example of a mammography exam is shown in Figure \ref{fig:mammo}. The dataset is annotated with breast-level cancer labels indicating biopsy-confirmed benign or malignant findings. Moreover, the dataset also provides bounding box annotations, and class labels (benign or malignant) of each visible positive findings. The entire dataset contains $985$ breasts with malignant findings and $5,556$ breasts with benign findings. The dataset is divided into training ($82\%$), validation ($5\%$) and test sets ($13\%$) ensuring a proportional distribution of benign and malignant cases across the subsets.

\begin{figure}[h!]
    \centering
    \subfloat[\centering]{{\includegraphics[width=0.3\linewidth]{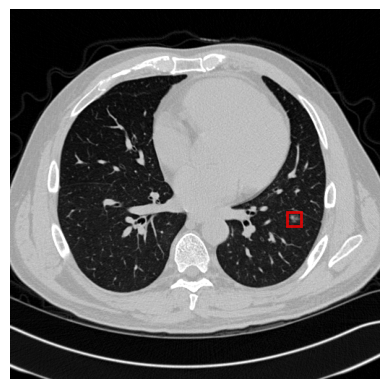} }}%
    \subfloat[\centering]{{\includegraphics[width=0.3\linewidth]{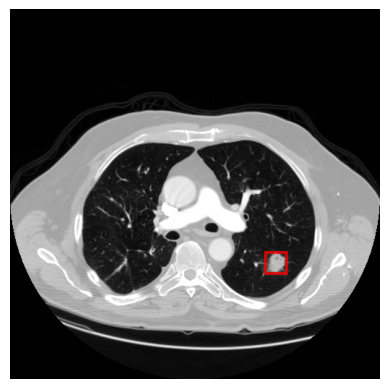} }}%
    \subfloat[\centering]{{\includegraphics[width=0.3\linewidth]{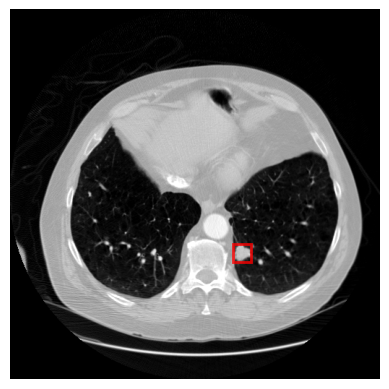} }}%
    \caption{\new{\textbf{2D CT scans with nodule annotations of LUNA16}} Each image shows a single axial CT slice with a red bounding box indicating the location of a lung nodule. These examples illustrate the small size and subtle appearance of nodules, highlighting the challenges of object detection in medical imaging.}%
    \label{fig:luna}%
\end{figure}

\newer{\paragraph{LUNA16} LUNA16~\citep{setio2017validation} is a public chest CT dataset for lung nodule detection, containing 888 3D chest CT scans with annotated nodule locations. We selected this dataset because it exemplifies key characteristics of medical imaging (Figure \ref{fig:luna}): (1) high resolution (typically 512 x 512 pixels per slice), necessary for capturing fine details; (2) small objects of interest, as nodules are subtle and occupy only a small portion of each scan; and (3) class imbalance, as nodules are relatively rare. Each nodule is annotated in 3D with a center point $(x,y,z)$ and diameter. To convert them into 2D bounding boxes, we identify the slices intersecting each nodule along the z-coordinate and project the center point to 2D $(x, y)$ coordinates on each slice. Using the diameter, we define a 2D bounding box around this point, allowing slice-by-slice nodule detection. Since DETR is designed for 2D object detection, we treat each 2D slice as an independent input to the model, enabling nodule detection in each slice separately. The dataset is randomly split for training (666 scans, 75\%), validation (88 scans, 10\%), and test (134 scans, 15\%).}

\subsection{Evaluation Metrics}
\label{sec:metrics}
In this study, we focus on evaluating the ability of the models to detect malignant lesions. We use Average Precision (AP) \citep{everingham2010pascal} and the Free-Response Receiver Operating Characteristic curve area (FAUC) \citep{bandos2009area}, which is a frequently used metric in medical image analysis \citep{yu2022systematic,wang2018development,petrick2013evaluation}. \new{Specifically, we focus on FAUC at the rate of 1 false positive per image or smaller, referred to as $\mathrm{FAUC}^1$, in line with the approach described by \citet{bandos2009area}. To formalize $\mathrm{FAUC}^1$, we introduce the following notation:}

\new{Let N denote the total number of images, indexed by $n=1,2,..,N$. Each image $n$ contains $L_n$ lesions with $L_{total}=\sum^N_1 L_n$ being the total lesions in the dataset. For each image $n$, a detection model produces a set of candidate detections $\mathcal{D}_n=\{d_n^1, d_n^2,...\}$, each with a confidence score $s(d_n^j)$. By varying a decision threshold $\tau$, one can include only those detections whose scores exceed $\tau$, denoted $\mathcal{D}_n(\tau) = \{d\in \mathcal{D}_n:s(d)\le \tau \}$.} 

Following prior works~\citep{article1, article2, 10.1001/jamanetworkopen.2023.0524}, we define a positive bounding box to have at least \new{10\%} Intersection over Union (IoU) with a ground truth box. These thresholds are deemed more appropriate for accurately detecting small-sized objects, such as cancerous lesions. \new{The $\mathrm{FAUC}^1_{10}$ metric integrates the true positive rate (TPR) over false positives per image (FPI) from 0 to 1: {\small$$\mathrm{FAUC}^1_{10} = \int^{1}_0\frac{1}{L_{total}} \sum^n_{n=1} \sum^{L_n}_{m=1} 1 ( \exists d\in \mathcal{D}_n(\tau(u)): d \leftrightarrow L_{n}^m) du, $$} where:
\begin{itemize}
    \item $\tau(u)$ is the threshold achieving $\mathrm{FPI} = \frac{1}{N}\sum_{N=1}^N|\{ d\in \mathcal{D}_n(\tau):  d \nleftrightarrow L_{n}^m, \forall m\in L_n \} = u$
    \item $d \leftrightarrow L_{n}^m$ indicates the lesion $m$ is detected by the prediction $d$ in image $n$
\end{itemize}} Following \new{the notation of integrated average precision in PASCAL VOC 2012 ~\citep{salton1983introduction, everingham2010pascal}}, we denote AP at 0.1 IoU threshold as $\mathrm{AP}_{10}$. Additionally, we report the average AP across IoU thresholds ranging from 0.1 to 0.5, in steps size of 0.05, denoted as $\mathrm{AP}_{10,50}$.

To clearly explain how well our models detect objects, we differentiate between ``localization'' and ``classification.'' 
\begin{itemize}
    \item \textbf{Localization} refers to the task of accurately drawing a bounding box around each ground-truth object. To be consistent with the definition of FAUC and AP, an object is considered successfully localized if the model produces a bounding box overlapping the ground truth box by more than 10\% IoU. To quantify localization accuracy, we compute the percentage of ground-truth objects successfully detected by the model. Assume there are $m$ ground-truth objects $G_i$ where $i=1,\ldots,m$ and $p$ predicted boxes $P_j$ where $j=1,\ldots,p$ in an image. The maximum IoU for a ground truth bound box $G_i$ among all predicted bounding boxes $P_{j}$ is $\max_j(\mathrm{IoU}(P_{j},G_i))$. Localization accuracy $L$ is then expressed as 
\begin{equation}\label{eq:l}
    L = \frac{1}{m} \sum^m_{i=1}\mathds{1}_{\max_j\left({\mathrm{IoU}\left(P_{j},G_i\right)}\right) \ge 0.1},
\end{equation}
\new{where the indicator function is defined as
{\small\[
\mathds{1}_{\max_j\left({\mathrm{IoU}\left(P_{j},G_i\right)}\right) \ge 0.1} = 
\begin{cases}
    1, & \mathrm{if } \max_j\left(\mathrm{IoU}\left(P_{j},G_i\right)\right) \ge 0.1 \\
    0, & \mathrm{otherwise}.
\end{cases}
\]}}

\item \textbf{Classification} involves associating the object inside each predicted box with the correct class. We consider models' classification accuracy using the percentage of successfully localized objects among the predicted bounding boxes with the top 10 highest predicted scores in each image. Among all the predicted bounding boxes $P_{j}$ in an image, let $S$ be the subset of indices of the top 10 predicted bounding boxes in an image. Localization performance considering classification is expressed as
\begin{equation}\label{eq:ltop10}
    L_\mathrm{top10} = \frac{1}{m} \sum^m_{i=1}\mathds{1}_{\max_{j \in S}({\mathrm{IoU}(P_j,G_i)}) \ge 0.1},
\end{equation}
where the indicator function is defined as
{\small\begin{align*}
\mathds{1}_{\max_{j \in S}\left({\mathrm{IoU}\left(P_{j},G_i\right)}\right) \ge 0.1}& = \\
&\begin{cases}
    1, & \mathrm{if } \max_{j \in S}\left(\mathrm{IoU}\left(P_{j},G_i\right)\right) \ge 0.1 \\
    0, & \mathrm{otherwise}.
\end{cases}
\end{align*}}

\end{itemize}

\subsection{Experimental Setup}
Our baseline model is Deformable DETR in its default setting, using a Swin-T backbone~\citep{liu2021swin}. For the NYU Breast Cancer Screening dataset, the backbone is pretrained on a breast cancer classification task using the same dataset (see Appendix~\ref{app:backbone} for details). Models are trained for 60 epochs on NYU Breast and 100 epochs on LUNA16. We used a batch size of 2 for NYU Breast and 32 for LUNA16. All models use the AdamW optimizer~\citep{loshchilov2017decoupled} with a step learning rate scheduler, which reduces the learning rate by a factor of 0.1 during the final 20 epochs. We tuned the hyperparameters using random search as detailed in Appendix \ref{app:tuning}. To account for training variability, we train five models with different random seeds for each experiment and report the mean and standard deviation of their performance. All training is conducted using a single NVIDIA A100 GPU.

\section{Results}
\label{sec:results}
Our experiments across the five design choices, including input resolutions, encoder layer complexity, multi-scale feature fusion, number of object queries, and two decoding techniques, reveal that standard Deformable DETR configurations do not align well with the unique characteristics of medical imaging datasets. This misalignment results in unnecessary computational overhead and sub-optimal performance.

\paragraph{Input Resolution}
Our experiments reveal a positive correlation between input resolution and detection performance, up to a certain point, for both the NYU Breast and LUNA16 datasets (Table \ref{table:resolution}). Specifically, increasing the resolution from 25\% to 50\% of the original image size significantly improves performance. On NYU Breast, this yields gains of 9.8\% in $\mathrm{FAUC}^1_{10}$, 8.6\% in $\mathrm{AP}_{10}$, and 6.4\% in $\mathrm{AP}_{10,50}$. Similarly, LUNA16 shows improvements of 5.9\%, 11.8\%, and 6.0\% in the corresponding metrics. Raising the resolution to $75\%$ continues to improve performance, although with diminishing returns in the NYU Breast. Interestingly, full-resolution images result in a decline in performance across all metrics on both datasets. This may be attributed to the limitations of the deformable attention mechanism. In high-resolution images, objects of interest may be distributed across a wider spatial area. The deformable attention mechanism only focuses on a selective set of keys centered around the reference points, which may miss necessary information in high resolution images. This phenomenon aligns with previous findings that question the assumption that higher resolution always improves performance~\citep{sabottke2020effect, thambawita2021impact,Richter_2021}.

It is also important to note the computational trade-off: increasing the input resolution from 25\% to 100\% results in a 10–15$\times$ increase in GFLOPs. To balance accuracy with computational efficiency, we used half-resolution images (50\%) for subsequent NYU Breast experiments and 75\% resolution for LUNA16, as these settings offer the best trade-off between performance and resource usage.

\begin{table*}[!h]
\caption{\textbf{Standard Deformable DETR performance using different input image resolutions.} The full resolution of images from NYU Breast is $2944 \times 1920$ and LUNA16 is $512 \times 512$. The detection performance is measured by AP and FAUC, as defined in Section \ref{sec:metrics}. GFLOPs are reported using one billion floating point operations per second as the unit.}
\label{table:resolution}
\begin{adjustbox}{width=0.9\linewidth,center}
\begin{tabular}{ cc|ccccc|cc }
\toprule
Dataset &Image &$\mathrm{FAUC}^1_{10}\new{\pm \mathrm{SD}}$ & $\mathrm{AP}_{10}\new{\pm \mathrm{SD}}$ &  $\mathrm{AP}_{10,50}\new{\pm \mathrm{SD}}$ &  $\new{L\pm \mathrm{SD}}$&$\new{L_{top10}\pm \mathrm{SD}}$& GFLOPs\\ 
&Resolution &&&&&\\
\midrule
 NYU&$1.0$  &$0.676\pm0.010$ & $0.671\pm0.005$ &$0.477\pm0.006$&\new{$0.925\pm0.004$}&\new{$0.843\pm0.003$}& $4367$\\
  &$0.75$  &$0.688\pm0.010$ & $0.683\pm0.007$ &$0.490\pm0.007$&\new{$0.930\pm0.004$}&\new{$0.862\pm0.005$}& $2448$\\
  &$0.5$  &$0.689\pm0.008$ & $0.669\pm0.012$ & $0.464\pm0.019$&\new{$0.920\pm0.005$}&\new{$0.856\pm0.009$}&$1102$\\
  &$0.25$ &$0.591\pm0.016$ & $0.583\pm0.007$ &$0.400\pm0.021$& \new{$0.897\pm0.003$}&\new{$0.799\pm0.01$}&$284$\\
  \midrule

  LUNA&$1.0$  &$0.521\pm0.011$ & $0.363\pm0.018$ &$0.221\pm0.015$&\new{$0.959\pm0.006$}&\new{$0.651\pm0.012$}& $3425$\\
  &$0.75$  &$0.537\pm0.007$ & $0.392\pm0.013$ &$0.295\pm0.005$&\new{$0.959\pm0.002$}&\new{$0.663\pm0.011$}& $1966$\\
  &$0.5$  &$0.488\pm0.003$ & $0.340\pm0.014$ & $0.221\pm0.019$&\new{$0.941\pm0.011$}&\new{$0.628\pm0.011$}&$986$\\
  &$0.25$ &$0.461\pm0.030$ & $0.304\pm0.011$ &$0.161\pm0.022$& \new{$0.951\pm0.014$}&\new{$0.604\pm0.018$}&$340$\\

 \bottomrule
\end{tabular}
\end{adjustbox}
\end{table*}

\paragraph{Encoder Complexity: Number of Encoder Layers} We investigated the effect of varying the number of encoder layers in Deformable DETR and evaluated whether the full encoder depth is necessary for medical imaging tasks. To ensure generalizability, we conducted experiments using two distinct backbones, ResNet50 and Swin-T.

On the NYU Breast dataset, for both backbones, reducing encoder layers from six to one or three results in comparable performance in all three detection metrics, while reducing GFLOPs by up to 40\% (Table~\ref{table:enc}). In particular, encoder-free models (0 layers) with Swin-T maintain performance within 1\% of the full 6-layer model, while cutting computation nearly half. This is likely because Swin-T was pretrained on the same mammography dataset, allowing it to extract strong task-specific features and reducing its reliance on the encoder. On LUNA16, we observed a similar pattern, with one or three encoder layers yielding a performance comparable to that of the full model, but the encoder-free models fail completely. This is likely due to not pre-training Swin-T backbone on lung CT, highlighting that when the backbone is not adapted to the target domain, some encoder capacity becomes necessary. Nevertheless, even with minimal encoder depth (one layer), the model achieved strong results while significantly lowering computational cost, from 1966 to 1225 GFLOPs.

These results suggest that the encoder can be shallower in DETR, regardless of whether the backbone is pretrained. When a powerful, domain-adapted backbone is available, the encoder can be removed with minimal impact on performance. This observation aligns with the recent development of the encoder-free $\mathrm{D}^2\mathrm{ETR}$~\citep{lin2022d}, which outperforms the standard DETR model on the MS COCO dataset \citep{lin2014microsoft}. Together, these insights challenge the conventional view that encoders are essential for feature transformation and multi-level feature integration within DETR models. Our results suggest that effective DETR-based detection can be achieved without encoders, particularly when paired with powerful backbones, offering a promising path toward more efficient and streamlined model designs.

\begin{table*}[!h]
\caption{\textbf{Varying the number of encoder layers in Deformable DETR.} The standard Deformable DETR has $6$ encoder layers. For both datasets, we do not observe any significant performance drop when using fewer encoder layers.}

\begin{adjustbox}{width=0.9\linewidth,center}

\label{table:enc}
\begin{tabular}{ ccc|ccccc|cc }
\toprule
Dataset&backbone & \#encoder & $\mathrm{FAUC}^1_{10}\new{\pm \mathrm{SD}}$ & $\mathrm{AP}_{10}\new{\pm \mathrm{SD}}$ &  $\mathrm{AP}_{10,50}\new{\pm \mathrm{SD}}$& $\new{L\pm \mathrm{SD}}$&$\new{L_{top10}\pm \mathrm{SD}}$&\#params & GFLOPs\\
&& layers & & & & & \\ 
\midrule
NYU&ResNet50 &$0$ & $0.643\pm0.020$& $0.619\pm0.010$ &$0.421\pm0.025$& \new{$0.910\pm0.013$} &\new{$0.819\pm0.010$}&$35.4$ & $536$ \\ 
&ResNet50& $1$ &   $0.656\pm0.011$&  $0.624\pm0.016$ &$0.439\pm0.005$& \new{$0.909\pm0.005$} &\new{$0.814\pm0.011$}&$36.2$ & $624$ \\ 
&ResNet50 &$3$ &$0.655\pm0.016$&  $0.627\pm0.013$ &$0.439\pm0.015$&\new{$0.910\pm0.009$}&\new{$0.818\pm0.014$}& $37.7$ & $801$\\ 
&ResNet50 &$6$ &$0.657\pm0.009$&$0.626\pm0.009$ &$0.436\pm0.011$&\new{$0.910\pm0.015$}&\new{$0.820\pm0.02$}&$40.0$ & $1067$\\  
\midrule
NYU&Swin-T &$0$  &$0.681\pm0.013$ & $0.662\pm0.013$ &$0.458\pm0.023$&\new{$0.923\pm0.006$}&\new{$0.843\pm0.008$}& $35.9$ & $570$\\
&Swin-T &$1$  &$0.684\pm0.009$ & $0.672\pm0.011$ &$0.463\pm0.014$& \new{$0.918\pm0.007$}&\new{$0.841\pm0.006$}&$36.7$ & $659$\\
&Swin-T &$3$  &$0.688\pm0.012$ & $0.677\pm0.011$ &$0.470\pm0.019$& \new{$0.918\pm0.005$}&\new{$0.855\pm0.011$}&$38.3$ & $836$\\
&Swin-T &$6$  &$0.689\pm0.008$ & $0.669\pm0.012$ & $0.464\pm0.019$&\new{$0.920\pm0.005$}&\new{$0.856\pm0.009$}&$40.5$ & $1102$\\
\midrule
LUNA&Swin-T &$0$  &$0.011\pm0.011$ & $0.001\pm0.000$ &$0.0\pm0.0$&\new{$0.433\pm0.085$}&\new{$0.089\pm0.022$}& $35.9$ & $1078$\\
&Swin-T &$1$  &$0.538\pm0.013$ & $0.390\pm0.005$ &$0.296\pm0.014$& \new{$0.966\pm0.002$}&\new{$0.680\pm0.016$}&$36.7$ & $1225$\\
&Swin-T &$3$  &$0.536\pm0.010$ & $0.399\pm0.003$ &$0.292\pm0.008$& \new{$0.966\pm0.008$}&\new{$0.672\pm0.008$}&$38.3$ & $1521$\\
&Swin-T &$6$  &$0.537\pm0.007$ & $0.392\pm0.013$ &$0.295\pm0.005$&\new{$0.959\pm0.002$}&\new{$0.663\pm0.011$}&40.5& $1966$\\
\bottomrule
\end{tabular}
\end{adjustbox}
\end{table*}

\paragraph{Encoder Complexity: Multi-Scale Feature Fusion}
 Standard Deformable DETR uses four feature maps of different scales in the encoder: three from the last three layers of the backbone and a fourth from a convolution applied to the backbone’s final output, (Figure~\ref{fig:DETR}(b)). Previous work show that multi-scale feature fusion improves detection performance on the MS COCO dataset as well as on other datasets~\citep{he2017mask,zhou2021mffenet,zeng2022small}. However, our results in Table \ref{table:feat} indicate that comparable performance can be achieved using only a single feature map of the backbone. This suggests that multi-scale feature fusion may not be necessary for detecting abnormalities in medical images.

\begin{figure}[h!]
    \centering
    \begin{minipage}{0.35\textwidth}
        \centering
    \includegraphics[width=\linewidth]{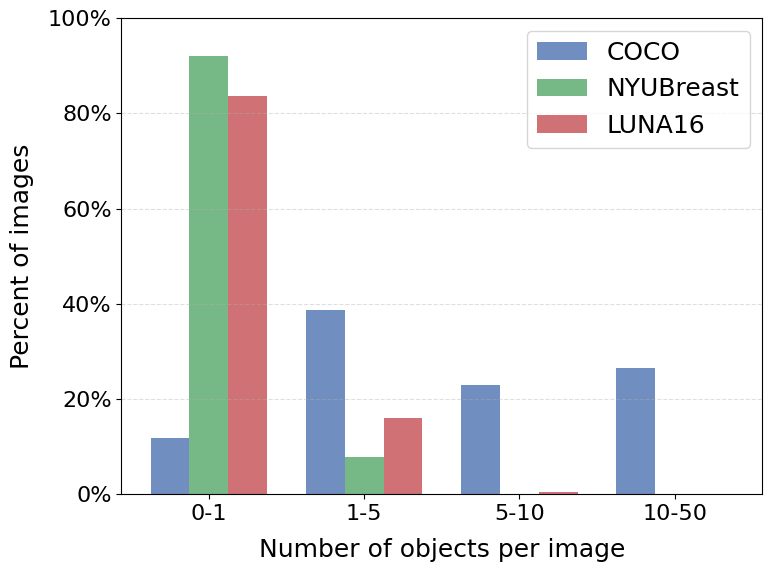} 
    \text{(a)}
    \end{minipage}
    \begin{minipage}{0.35\textwidth}
        \centering
    \includegraphics[width=\linewidth]{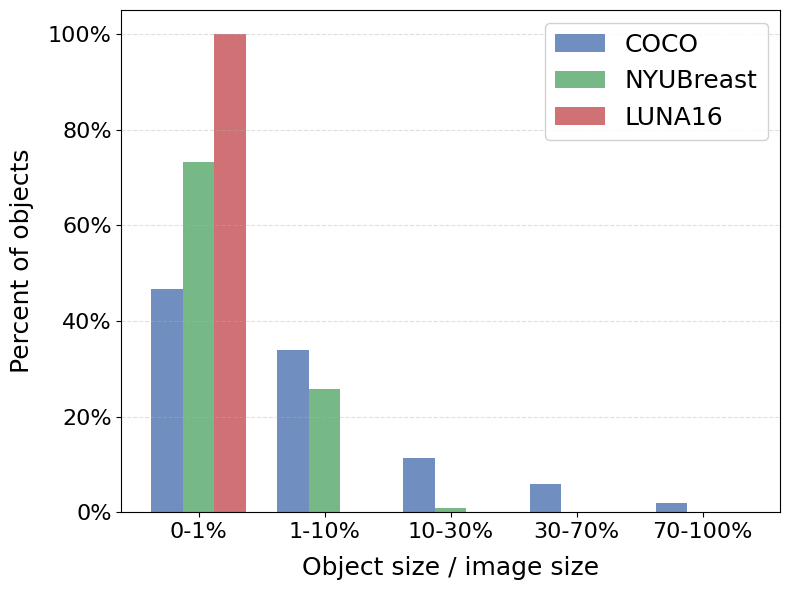}
    \text{(b)}
    \end{minipage}
    \caption{\textbf{Comparison of object count per image and object size variability across MS COCO, NYU Breast, and LUNA16 datasets.} In contrast to the MS COCO dataset, the medical imaging datasets, NYU Breast and LUNA16, contain fewer objects per image and show significantly less variation in object size. The standard deviation of normalized object sizes is 0.161 for COCO, 0.025 for NYU Breast, and 0.001 for LUNA16.}
    \label{fig:object-size}
\end{figure}

The characteristics of medical datasets likely explain this finding. The objects in natural image datasets, such as MS COCO, show a high variability in scale and quantity due to perspective, camera distance, and the inherent size differences between object classes (Figure~\ref{fig:image-examples}). Multi-scale fusion benefits such settings by enabling the model to attend to features at different resolutions, capturing objects of varying sizes more effectively. However, in medical datasets like NYU Breast and LUNA16, most images contain a single object and the sizes of these objects are relatively uniform (Figure \ref{fig:object-size}). This contrasts to the MS COCO dataset, showing a broader variation in both the sizes of objects and the number of objects per image. For such medical datasets, the benefits of multi-scale feature fusion are less pronounced. Consequently, in a homogeneous dataset, the additional complexity of multi-scale feature fusion may not translate into better performance. 

Notably, in LUNA16, we observed that using the last feature level resulted in a performance drop. This is likely due to the extremely small object size in LUNA16, where all nodules occupy on average 0.6\% of the image area (Figure \ref{fig:object-size} (b)). The last-layer feature map has too low a spatial resolution (i.e. downsized to $16\times16$) to preserve the fine-grained detail necessary for detecting such small objects. This highlights that while multi-scale fusion may not be generally required in medical imaging, selecting an appropriate single feature level, especially one with sufficient spatial resolution, is still critical for detecting very small targets.

\begin{table*}[!ht]
\caption{\textbf{The performance of Deformable DETRs using different combinations of feature levels.} The standard Deformable DETR uses all 4 levels of feature maps from the backbone. Using only the 3rd level feature map for breast NYU and 2nd level feature map for LUNA16, achieves on-par with or even better performance than using multi-level feature fusion.}
\label{table:feat}
\begin{adjustbox}{width=0.9\linewidth,center}
\begin{tabular}{ cc|ccccc|cc }
\toprule
 Dataset&Feature Levels & $\mathrm{FAUC}^1_{10}\new{\pm \mathrm{SD}}$ & $\mathrm{AP}_{10}\new{\pm \mathrm{SD}}$ &  $\mathrm{AP}_{10,50}\new{\pm \mathrm{SD}}$& $\new{L\pm \mathrm{SD}}$&$\new{L_{top10}\pm \mathrm{SD}}$&\# params & GFLOPs\\ 
\midrule
NYU&$1,2,3,4$ (standard) &$0.688\pm0.012$ & $0.677\pm0.011$ &$0.470\pm0.019$ & \new{$0.918\pm0.005$}&\new{$0.855\pm0.011$}&$38.3$ & $836$ \\ 
&$1$ & $0.637\pm0.012$ & $0.627\pm0.009$ &$0.434\pm 0.009$ & \new{$0.924\pm0.007$}&\new{$0.849\pm0.016$}&$35.5$& $734$  \\ 
&$2$ &  $0.680\pm0.005$ & $0.666\pm0.009$ &$0.467\pm0.016$ & \new{$0.917\pm0.005$}&\new{$0.851\pm0.012$}&$35.6$& $570$\\ 
&$3$ &  $0.688\pm0.004$ & $0.675\pm0.007$ &$0.475\pm0.006$ & \new{$0.915\pm0.010$}&\new{$0.858\pm0.009$}&$35.7$& $528$\\  
\midrule
LUNA&$1,2,3,4$ (standard) &$0.538\pm0.013$ & $0.390\pm0.005$ &$0.296\pm0.014$& \new{$0.966\pm0.002$}&\new{$0.680\pm0.016$}&$36.7$ & $1225$\\
&$1$ & $0.540\pm0.007$ & $0.378\pm0.022$ &$0.267\pm0.020$ & \new{$0.961\pm0.004$}&\new{$0.674\pm0.022$}&$34.2$& $1141$\\ 
&$2$ &  $0.548\pm0.017$ & $0.403\pm0.006$ &$0.309\pm 0.008$ & \new{$0.965\pm0.003$}&\new{$0.668\pm0.007$}&$34.3$& $1018$  \\ 
&$3$ &  $0.491\pm0.015$ & $0.290\pm0.025$ &$0.142\pm0.019$ & \new{$0.959\pm0.005$}&\new{$0.637\pm0.010$}&$34.4$& $987$\\  
\bottomrule
\end{tabular}
\end{adjustbox}
\end{table*}

\begin{figure}[h]
    \centering
    \subfloat[\centering $\mathrm{FAUC}^1_{10}$]{{\includegraphics[width=0.45\linewidth]{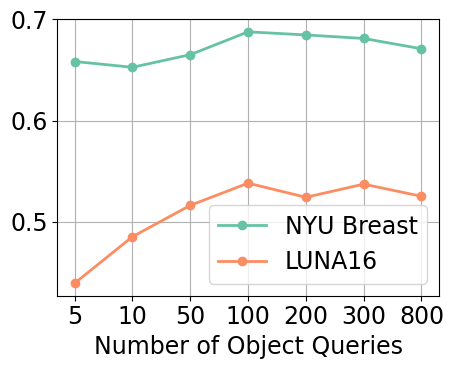} }}%
    \hspace{.001\textwidth}
    \subfloat[\centering $\mathrm{AP}_{10}$]{{\includegraphics[width=0.45\linewidth]{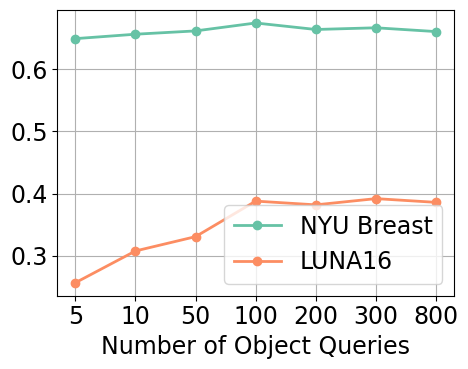} }}%
    \hspace{.001\textwidth}
    \subfloat[\centering $\mathrm{AP}_{10,50}$]{{\includegraphics[width=0.45\linewidth]{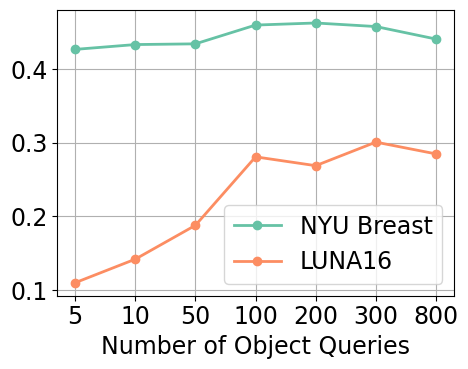} }}
    \subfloat[\centering $L$ and $L_{top10}$]{{\includegraphics[width=0.45\linewidth]{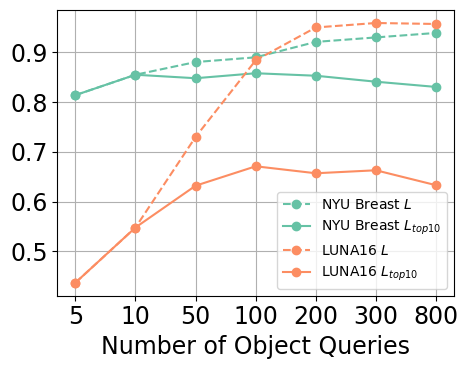} }}
    
\caption{\textbf{(a–c)} Detection performance of Deformable DETR across varying numbers of object queries on the NYU Breast and LUNA16 datasets. Performance improves as the number of queries increases from 5 to 100 but hit plateau when the number exceeds 100. \textbf{(d)} Localization performance ($L$) continues to improve with more queries, while classification performance ($L_{\mathrm{top10}}$) drops beyond 100 queries, suggesting an increase in false positives that displace true positives in the top-ranked predictions.}
\label{fig:query}
\end{figure}

\paragraph{Number of object queries} Figure \ref{fig:query}(a)-(c) illustrates the impact of increasing the number of object queries from 5 to 800 on detection performance across both the NYU Breast and LUNA16 datasets. Increasing the number of object queries from $5$ to $100$ consistently improves the detection performance. However, further increasing the number of queries beyond 100 results in diminishing returns and even a slight decline in performance. This pattern is consistent on both datasets, although more obvious on LUNA16.

To better understand this behavior, we examined the performance on localization $L$ (cf. Equation \ref{eq:l}) and classification $L_{\mathrm{top10}}$ (cf. Equation \ref{eq:ltop10}) separately in Figure~\ref{fig:query}(d). Localization performance ($L$) continues to improve with more object queries, indicating an enhanced ability to correctly localize objects. However, the classification performance ($L_{\mathrm{top10}}$), which measures how many correctly localized boxes rank among the top 10 predictions by classification score, declines beyond 100 queries. This suggests that while more queries increase the likelihood of finding true objects, they also introduce additional false positives that dilute the ranking of true positives.

We hypothesize that having more object queries increases the chances of localizing false positives. More object queries expand the model's search space, making it more sensitive to subtle features or noise that resemble the characteristics of true objects. This can lead to more false positives being assigned high classification scores, pushing true positives lower in the ranked predictions. This issue is especially relevant in medical imaging, where images typically contain only one or very few objects of interest. In such sparse-object settings, the increased false positive rate from excessive queries can outweigh the benefits of improved localization.

\paragraph{Decoding Techniques}
We evaluated two widely used decoding techniques in the DETR family, query initialization and iterative bounding box refinement (IBBR), using a simplified model with design choices achieved through previous results. As shown in Table~\ref{table:query-refinement}, neither technique significantly improved detection performance across $\mathrm{FAUC}^1_{10}$, $\mathrm{AP}_{10}$, or $\mathrm{AP}_{10,50}$ for both datasets.

\begin{figure*}[h!]
    \centering
    \subfloat[\centering NYU Breast]
    {\includegraphics[width=0.9\linewidth]{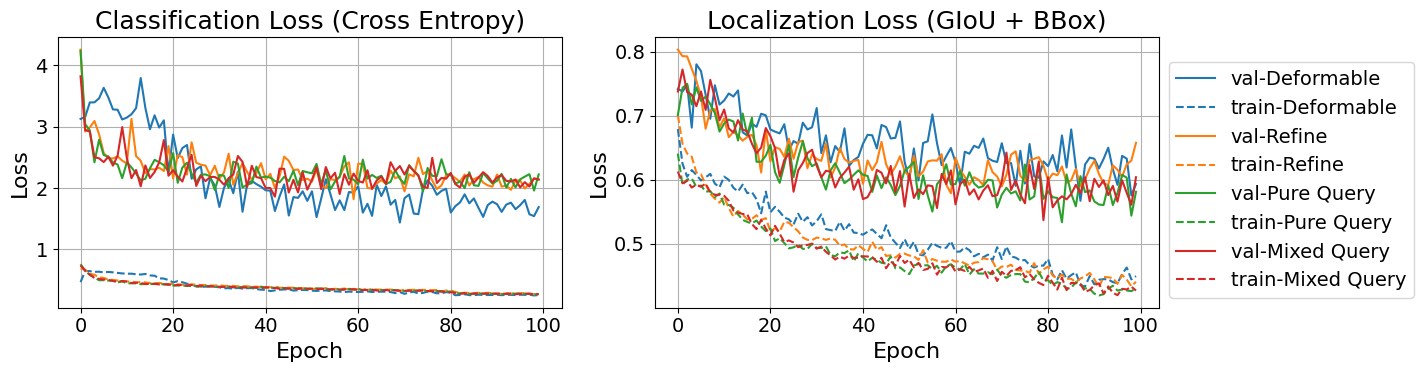} }%
    \hspace{.001\textwidth}
    \subfloat[\centering LUNA16]
    {\includegraphics[width=0.9\linewidth]{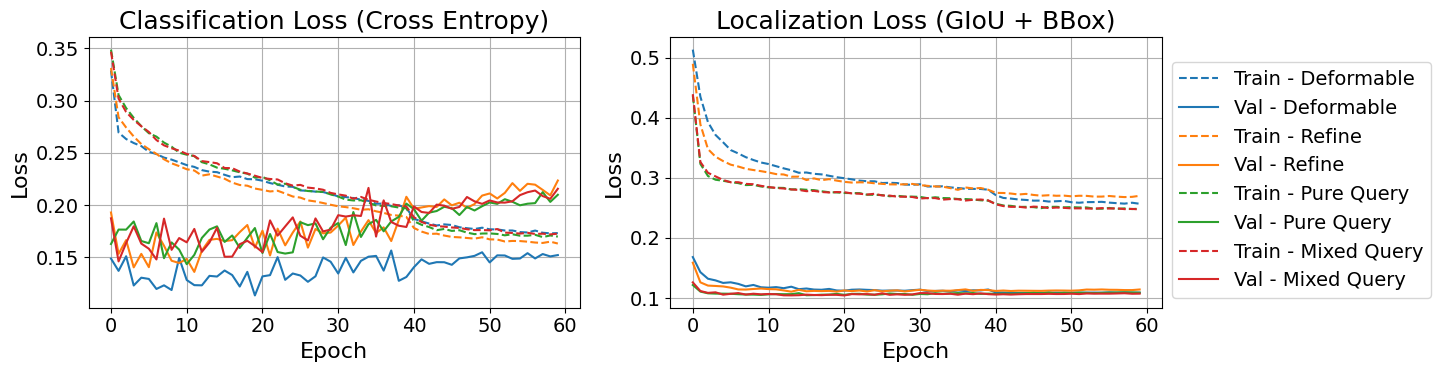} } 
\caption{\textbf{Training and validation losses of Deformable DETR models with and without decoding techniques for (a) NYU Breast dataset and (b) LUNA16 dataset.}
Each plot shows the average loss over five runs with different random seeds. The left panels display classification loss (binary cross-entropy), and the right panels show localization loss (GIoU + bounding box regression).  Models without decoding techniques (blue lines) consistently show less overfitting, especially in classification loss, compared to models using query initialization or iterative bounding box refinement.}
    \label{fig:refine}
\end{figure*}

To better understand this outcome, we separately analyzed localization and classification performance using $L$ and $L_{\mathrm{top10}}$. We found that while these techniques improved localization performance, they adversely affected classification performance. Figure~\ref{fig:refine}  shows training and validation losses for localization (IoU and box regression) and classification (binary cross-entropy). Models equipped with query initialization or IBBR display stronger overfitting, especially in classification loss, compared to the baseline model without these techniques. We hypothesize that this overfitting is due to the limited number of positive objects in our datasets. As the model becomes more effective at localizing regions of interest, it may focus too narrowly on those few positive examples, leading to memorization rather than learning generalizable features. This reduces the model’s ability to accurately distinguish between subtle classes, ultimately weakening the classification performance.

\begin{table*}[!h]
\caption{\textbf{Impact of query initialization methods and iterative bounding box refinement (IBBR) on Deformable DETR performance.} Detection performance is evaluated on models with the best configuration based on previous experiments for each dataset. Neither query initialization strategies (static, pure, or mixed) nor IBBR consistently improve performance across the three main detection metrics ($\mathrm{FAUC}^1_{10}$, $\mathrm{AP}_{10}$, $\mathrm{AP}_{10,50}$). While some configurations improve localization ($L$), classification performance ($L_{\mathrm{top10}}$) generally declines.}

\label{table:query-refinement}
\begin{adjustbox}{width=0.9\linewidth,center}
\begin{tabular}{ ccc|ccccc }
\toprule
Dataset&Refinement &Query Initial. & $\mathrm{FAUC}^1_{10}\new{\pm \mathrm{SD}}$ & $\mathrm{AP}_{10}\new{\pm \mathrm{SD}}$ &  $\mathrm{AP}_{10,50}\new{\pm \mathrm{SD}}$&  $\new{L\pm \mathrm{SD}}$&$\new{L_{top10}\pm \mathrm{SD}}$ \\ 
\midrule
NYU&&Static & $0.688\pm0.004$ & $0.675\pm0.011$ &$0.475\pm 0.019$ & \new{$0.915\pm0.010$}&\new{$0.858\pm0.009$}\\ 
&&Pure& $0.678\pm0.007$&	$0.668\pm0.011$&$0.474\pm0.018$& \new{$0.928\pm0.011$}&\new{$0.833\pm0.015$}\\ 
&&Mixed& $0.670\pm0.012$&$0.633\pm0.010$&$0.443\pm 0.014$  & \new{$0.936\pm0.005$}&\new{$0.834\pm0.012$}\\ 
 &$\checkmark$ &Static & $0.680\pm 0.010$ &$0.678\pm0.012$&$0.470\pm0.009$  & \new{$0.931\pm0.003$}&\new{$0.848\pm0.009$}\\ 
&$\checkmark$ &Pure & $0.670\pm0.010$& $0.670\pm0.019$&$0.468\pm 0.013$ & \new{$0.957\pm0.008$}&\new{$0.838\pm0.007$} \\
\midrule
LUNA&& Static &  $0.548\pm0.017$ & $0.403\pm0.006$ &$0.309\pm 0.008$ &\new{$0.965\pm0.003$}&\new{$0.668\pm0.007$}\\ 
&&Pure& $0.520\pm0.007$&$0.396\pm0.006$&$0.298\pm0.008$& \new{$0.965\pm0.006$}&\new{$0.651\pm0.008$}\\ 
&&Mixed& $0.500\pm0.018$&$0.383\pm0.007$&$0.291\pm 0.013$  & \new{$0.967\pm0.006$}&\new{$0.636\pm0.022$}\\ 
 &$\checkmark$ &Static & $0.516\pm 0.013$ &$0.393\pm0.017$&$0.303\pm0.022$  & \new{$0.969\pm0.014$}&\new{$0.639\pm0.004$}\\ 
&$\checkmark$ &Pure & $0.521\pm0.015$& $0.399\pm0.003$&$0.300\pm 0.003$ & \new{$0.965\pm0.007$}&\new{$0.635\pm0.005$} \\
\bottomrule
\end{tabular}
\end{adjustbox}
\end{table*}

\begin{figure}[h!]
    \centering
    \subfloat[\centering $0.814$]{{\includegraphics[width=0.45\linewidth]{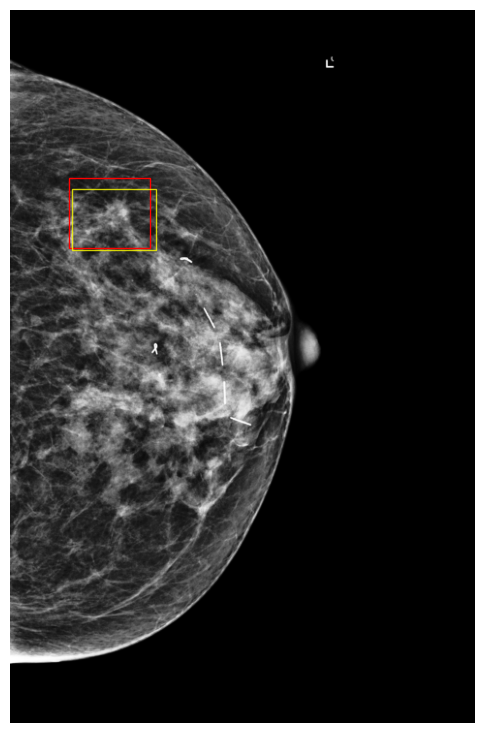} }}%
    \hspace{.005\textwidth}
    \subfloat[\centering $0.823$]{{\includegraphics[width=0.45\linewidth]{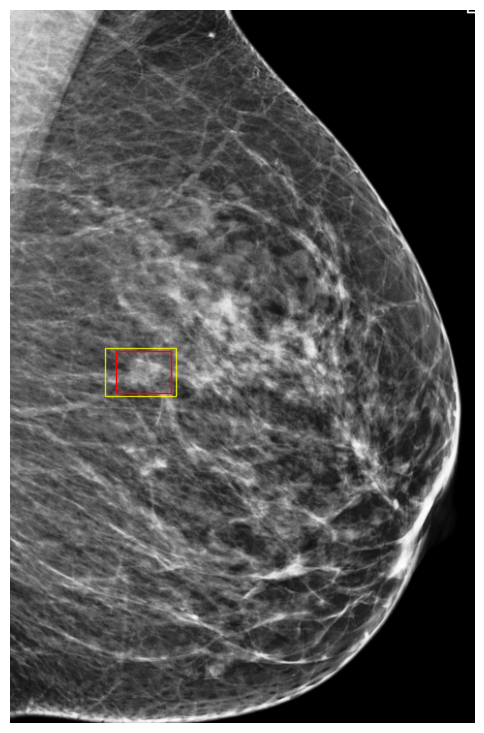} }}%
    \hspace{.005\textwidth}
    \subfloat[\centering $0.825$]{{\includegraphics[width=0.45\linewidth]{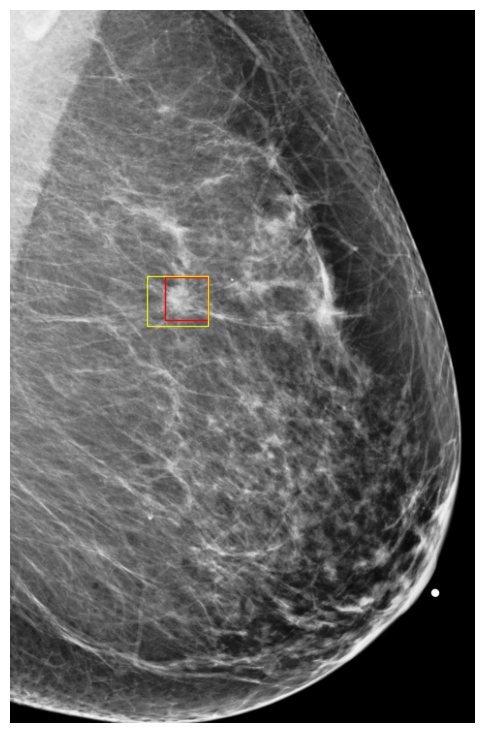} }}%
    \hspace{.005\textwidth}
    \subfloat[\centering $0.822$]{{\includegraphics[width=0.45\linewidth]{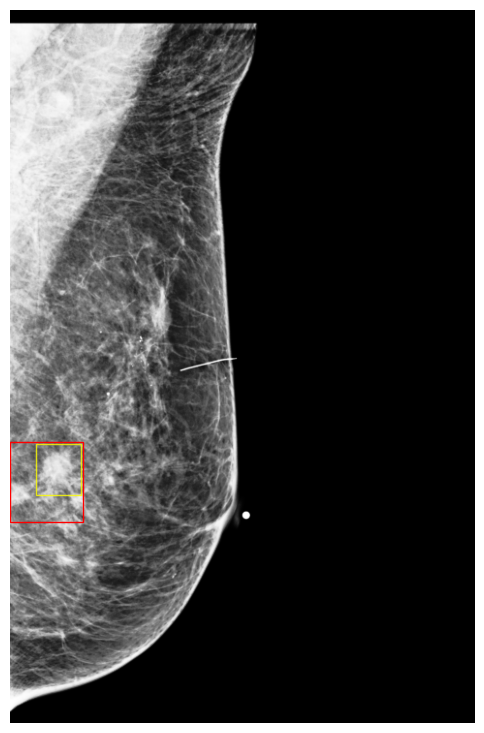} }}%
    \caption{\textbf{Example mammograms with high classification scores} They tend to be higher density masses with non-circumscribed, irregular or indistinct borders, strongly suggestive of malignancy. The red bounding boxes are ground truth annotations and the yellow bounding boxes are the prediction of our model. The prediction results are provided by Deformable DETR with pure query initialization method and IBBF, as detailed in Table~\ref{table:query-refinement}.}%
    \label{fig:mammo-high}%
\end{figure}

\begin{figure}[h!]
    \centering
    \subfloat[\centering $0.014$]{{\includegraphics[width=0.45\linewidth]{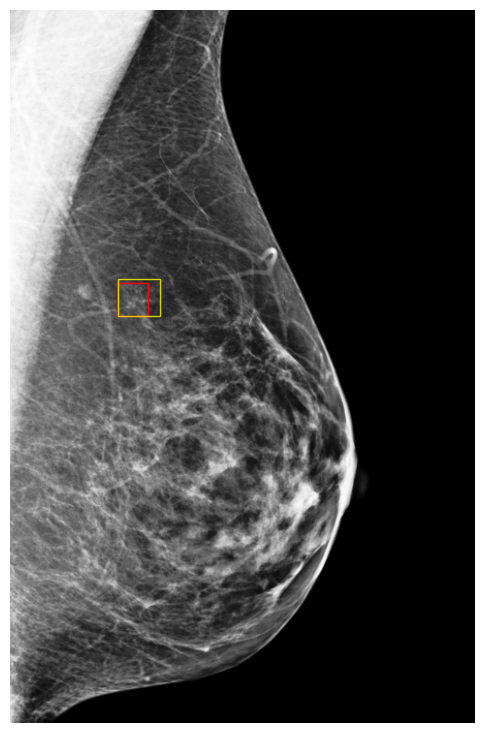} }}%
    \hspace{.005\textwidth}
    \subfloat[\centering $0.018$]{{\includegraphics[width=0.45\linewidth]{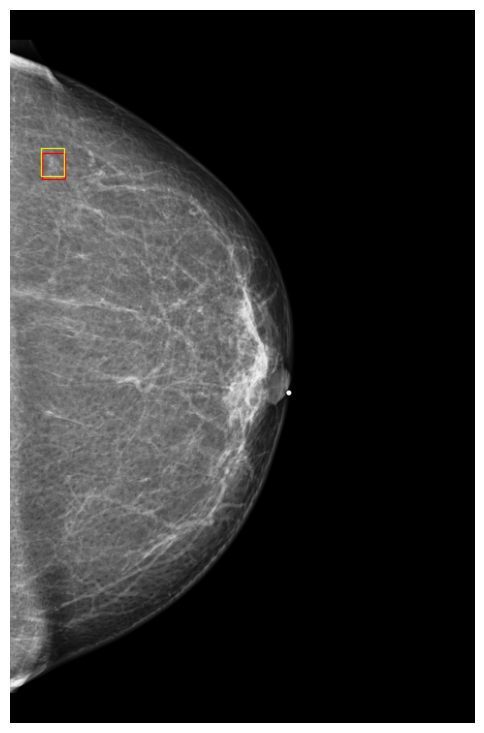} }}%
    \hspace{.005\textwidth}
    \subfloat[\centering $0.040$]{{\includegraphics[width=0.45\linewidth]{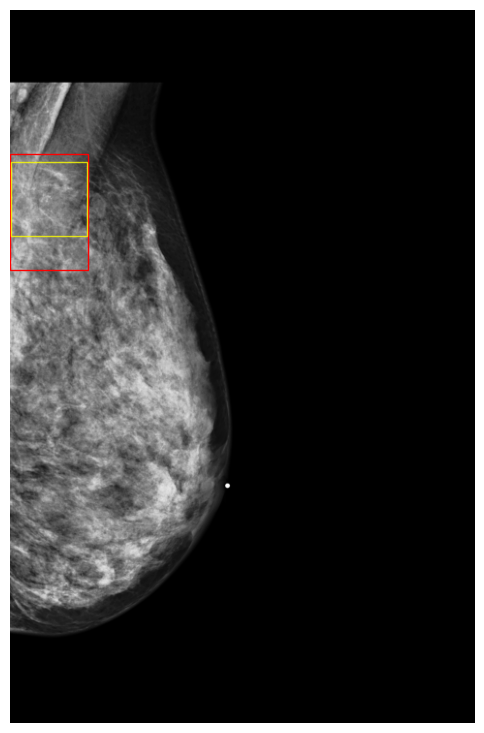} }}%
    \hspace{.005\textwidth}
    \subfloat[\centering $0.044$]{{\includegraphics[width=0.45\linewidth]{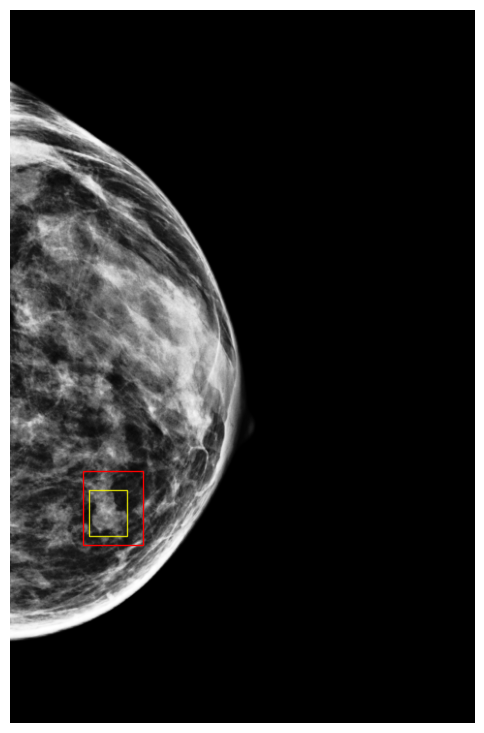} }}%
    \caption{\textbf{Example mammograms with lowest classification scores} They are mostly asymmetric tissue or low density masses with circumscribed borders, more likely to be false-positives. The red bounding boxes are ground truth annotations and the yellow bounding boxes are the prediction of our model. The prediction results are provided by Deformable DETR with pure query initialization method and IBBF, as detailed in Table~\ref{table:query-refinement}.}%
    \label{fig:mammo-low}%
\end{figure}

\paragraph{Cases Visualization on NYU Breast} 
Finally, to better understand how DETR models make predictions, we visualized a few exams along with their classification scores on NYU Breast dataset. Figures \ref{fig:mammo-high} and \ref{fig:mammo-low} show images in which the model assigned cancerous objects high malignant scores (scores $\ge 0.8$) and low scores (scores $\le 0.1$), respectively. We observed that the model correctly localizes abnormal objects in all images. However, it tends to assign high scores to high-density masses featuring non-circumscribed, irregular, or indistinct borders, which are typically indicative of malignancy to the human eye. In contrast, the model usually assigns low scores to low-density masses with circumscribed borders, which can easily be confused with benign cases \citep{10.1093/med/9780190270261.003.0024}.

\section{Conclusions}
In this study, we investigated the impact of common design choices in Deformable DETR\new{~\citep{zhu2020deformable}} on object detection performance in medical imaging, using two representative datasets: the NYU Breast Cancer Screening Dataset and LUNA16. We found that all the design choices we experimented with need to be reconsidered, and simpler architectures typically lead to better performance on medical dataset.

Additionally, our findings suggest that the model tends to struggle more with correctly classifying detected objects than with localizing them. Many design choices developed for natural image detection, such as query initialization, multi-scale feature fusion, and bounding box refinement, are primarily aimed at improving localization. However, since classification appears to be the more challenging component in medical imaging, these localization-focused techniques may offer limited benefit and, in some cases, even hinder performance.


Future research should focus on developing architectures specifically tailored to the characteristics of medical imaging. This includes improving the model’s ability to extract subtle visual cues that are often critical for classification, such as texture variations, tissue density changes, irregular borders, and microcalcifications. Another important direction is designing architectures that can efficiently process full-resolution images, allowing the model to leverage detailed information in relevant regions while minimizing the influence of background areas. Moreover, addressing overfitting in classification tasks, particularly in datasets with limited positive samples, requires the integration of effective regularization techniques to improve generalization and robustness.

\section{Limitations and future work}
Our study has several limitations. First, while our results demonstrate that simplified DETR configurations perform well on medical imaging tasks, future studies should explore additional architectural designs within the DETR family to validate and extend these findings, for example, constrastive denoising training in DINO \citep{zhang2022dino} and anchors in Anchor-DETR \citep{wang2022anchor}. Second, our experiments were conducted primarily on the NYU Breast Cancer Screening Dataset and LUNA16 for lung nodule detection. While these datasets capture important aspects of medical imaging, future studies should evaluate model performance across a broader set of imaging modalities and clinical tasks, such as brain MRI, ultrasound, or multi-phase CT, to assess the generalizability of our conclusions.

\acks{This work was supported in part by grants from the National Institutes of Health (P41EB017183), the National Science Foundation (1922658), the Gordon and Betty Moore Foundation (9683), and the Mary Kay Ash Foundation (05-22). We also appreciate the support of Nvidia Corporation with the donation of some of the GPUs used in this research.}

%
\ethics{\new{This retrospective study was approved by the NYU Langone Health Institutional Review Board (ID\#i18-00712\_CR3) and is compliant with the Health Insurance Portability and Accountability Act. Informed consent was waived since the study presents no more than minimal risk.}}

\coi{The authors do not declare any conflicts of interest.}

\data{Our internal (NYU Langone Health) dataset is not publicly available due to internal data transfer policies. We released a data report on data curation and preprocessing to encourage reproducibility. The data report can be accessed at \href{https://cs.nyu.edu/~kgeras/reports/datav1.0.pdf}{this link}.}

\bibliography{main}

\clearpage
\appendix

\section{DETR architecture}
\label{app:architecture}

\subsection{Multi-head self-attention (MHSA)} 
\label{app:mhsa}
A standard MHSA with $M$ heads is defined as: 
{\small\begin{align}\label{eq:1}
    & \mathrm{MHSA}(Q,K,V) \\
    &= \sum_{m=1}^M W_{mo}\left[\mathrm{softmax}\left(\frac{QW_{mq}(KW_{mk})^T}{\sqrt{d/M}}\right)VW_{mv}\right].
\end{align}}
The $K$, $Q$ and $V$ represent the query, key, and value matrices respectively, defined with respect to input feature map $x_f\in \mathbb{R}^{WH \times d}$ and its positional embedding $x_p\in \mathbb{R}^{WH \times d}$:
\begin{equation}
    Q = x_f + x_p,\space K=x_f+x_p, \space V = x_f.
\end{equation}
$W_{mq}, W_{mk}, W_{mv} \in \mathbb{R}^{d\times d/M}$ linearly transforms $K,Q,V$ in the $m$-th head and $W_{mo}\in \mathbb{R}^{d/M\times d }$.

\subsection{Multi-head (MH) cross-attention} 
\label{app:mh-cross}
The MH cross-attention module performs the same computation as the MHSA defined in \ref{eq:1}, except that $K,Q,V$ are defined based on two different sets of tokens.  The queries $Q$ are defined by the object queries $q = q_c+q_p$, where $q_p$ and $q_c$ are the positional embedding and the content embedding of the object queries. The keys $K$ are defined by the encoder features $x_{enc} + x_p$. Specifically,
\begin{equation}
    Q = q_c + q_p,\space K=x_{enc}+x_p, \space V = q_c.
\end{equation}

\subsection{Set prediction loss}
\label{app:loss}

DETR uses a set prediction loss that enables end-to-end training without non-maximum suppression (NMS). DETR produces a fixed number of predictions per image $N$, and $N$ is set to be significantly larger than the maximum possible number of objects in the image. Let $\{\hat{y}_i = (\hat{c}_i, \hat{b}_i)\}_N$ be all pairs of class and box predictions. The set of $N$ labels is $\{y_i = (c_i, b_i )\}_N$ where each ground truth label represents an object in the image. If there are fewer objects than $N$, the rest of the labels are empty classes $(0, \emptyset)$. The set prediction loss is computed in two steps. The first step is to find a permutation $\sigma$ on the set of labels $\{y_i\}$ that minimizes the matching loss, defined as below: $$
\mathbf{\hat{\sigma}} = \arg \min_{\sigma\in \Sigma_N} \sum_N^i \mathcal{L}_{match}(\hat{y}_i,y_{\sigma(i)}).$$
The matching loss for a matching pair is a linear combination of classification loss, box regression loss and GIoU loss \citep{Rezatofighi_2018_CVPR}. The classification loss is a standard focal loss \citep{lin2017focal}. The regression loss and the GIoU loss are only applied to non-empty labels. It is defined as the following: 

$$\mathcal{L}_{match}(\hat{y}_i,y_{\sigma(i)}) = W_{cls} \mathcal{L}_{cls}(\hat{c}_i,c_{\sigma(i)}) $$
$$+ 1_{b \neq \emptyset} \left(W_{l1}\mathcal{L}_{l1}(\hat{b}_i,b_{\sigma(i)}) + W_{giou} \mathcal{L}_{giou}(\hat{b}_i,b_{\sigma(i)})\right),$$
where $W_{cls}, W_{l1}, W_{giou}$ are scalar coefficients that are tuned as hyperparameters to balance the scale of different losses. The Hungarian algorithm \citep{kuhn1955hungarian} can efficiently find the optimal match $\hat{\sigma}$. The second step is to minimize the loss function $\sum_N^i \mathcal{L}_{match}(\hat{y}_i,y_{\hat{\sigma}(i)})$ with the permutation $\hat{\sigma}$ on the label set.

DETR also utilizes auxiliary loss in each decoder layer to provide stronger supervision. At the end of each decoder layer, it predicts $N$ boxes and class scores with MLP prediction heads. All prediction heads share weights. The above two steps, the matching step and the Hungarian loss minimization, are applied to each decoder layer's output. In inference, only the output of the last layer is used as the final prediction.

\section{Deformable DETR architecture}
\subsection{Deformable multi-head self-attention}
\label{app:deformable-mhsa}
Formally, deformable MHSA for a single query $q \in \mathbb{R}^{d}$ in the feature map is given by:
\begin{align}
    &\mathrm{Deform\_MHSA}(Q_q,K,V) \\
    &= \sum_{m=1}^M W_{mo} \left[\mathrm{softmax}\left(K_{q}W_{mk}\right)V_qW_{mv}\right] 
\end{align}
The $Q$, $K$ and $V$ represent the query, key, and value matrices respectively, defined as the following,
\begin{align}
    &Q = x_f + x_p, \\
    &Q_q = q, \\
    &K_{q}= \delta(K,q)\in \mathbb{R}^{k\times d}, \\
    &V_q = \delta(V,q) \in \mathbb{R}^{k\times d}.
\end{align}
The key sampling function $\delta$ samples the $k$ keys from the full set of keys $K= x_f + x_p$ by generating the sampling offsets $\Delta p$ with respect to reference points $p_q$: $\delta(K,q) = K(p_q + \Delta p)$. The sampling offsets are obtained by linear transformation of the query $q$.

\section{Iterative Bounding Box Refinement Technique}
\label{app:refine}
In the standard Deformable DETR, a 2D reference point $r_q\in [0,1]^{2}$ for each object query $q$ is derived from its learnable positional embedding $p_q$ via a linear layer $$ r_q = \mathrm{linear}(p_q).$$ Throughout the decoder, the locations of these reference points remain constant. They are updated based on the learnable positional embedding $p_q$ when a backward pass is completed. Formally, let $r_q^i$ be the reference points of an object query $q$ in the i-th decoder layer. In standard Deformable DETR, $$ r_q^1 = r_q^2 = \ldots = r_q^6.$$ In IBBR, the reference points $r_q^i$ in i-th decoder layer are refined based on the previous reference points $r_q^{i-1}$ and the offsets predicted by the auxiliary prediction head, \new{which is a Multi-layer Perceptron (MLP). MLP is defined by three fully connected layers, which transform the output embeddings of the transformer into the desired bounding box coordinates.} $$r_q^{i} = S(\mathrm{linear}^{-1}(S^{-1}(r_q^{i-1})) + \mathrm{MLP}(x_{dec}^i)),$$ where $S$ and $S^{-1}$ represent the sigmoid function and its inverse, and $x^i_{dec}$ is the output of the i-th decoder layer.
\section{Backbone Pre-training}
\label{app:backbone}

We pretrained the Swin-T Transfromer backbone with a cancer classification task on our dataset. This classification task is a binary multi-label classification that predicts two scores indicating if an input image contains benign lesions \new{and/or} malignant lesions.

\section{Hyperparameter Tuning}
\label{app:tuning}
Our method for hyper-parameter tuning is random search. We tuned the following hyperparameters and their ranges on quarter-resolution images:
\begin{itemize}
\item learning rate $\eta \in 10^{[3,5.5]}$,
\item scale of the backbone learning rate $s\in [1, 0.01]$ (backbone learning rate = $s\times \eta$ ),
\item weight decay $\lambda \in 10^{[3,6]}$,
\item number of object queries $N \in [10, 200]$,
\item two hyperparameters $\alpha$ and $\gamma$ in the focal loss $\alpha \in [0,1]$, $\gamma\in [0,3]$,
\item the coefficients on classification loss and GIoU loss $\in [0,1]$.
\end{itemize} 
We train $80$ jobs in total and choose the best model based on \new{$\mathrm{FAUC}^1_{10}$}. 

\end{document}